%% file: main.tex
\documentclass[a4paper, 10pt, conference]{ieeeconf}      
\usepackage{FG2024}

\FGfinalcopy 

\IEEEoverridecommandlockouts                              
\usepackage{graphicx}
\usepackage{amsmath}
\usepackage{amssymb}
\usepackage{booktabs}
\usepackage{bbm}
\usepackage{xspace}
\usepackage{url}

\def\FGPaperID{41} 

\title{\LARGE \bf
Geometry-Biased Transformer for Robust Multi-View 3D Human Pose Reconstruction
}

\author{\parbox{16cm}{\centering
    {\large Olivier Moliner$^{1,2}$, Sangxia Huang$^{2}$ and Kalle Åström$^{1}$}\\
    {\normalsize
    $^1$ Centre for Mathematical Sciences, Lund University, Sweden\\
    $^2$ Sony Corporation, Lund Laboratory, Sweden}}
    \thanks{This work was partially supported by the Wallenberg AI, Autonomous Systems and Software Program (WASP) funded by the Knut and Alice Wallenberg Foundation.}
}

\newcommand\T{\rule{0pt}{2.6ex}}       
\newcommand\B{\rule[-1.2ex]{0pt}{0pt}} 

\usepackage[capitalize]{cleveref}
\crefname{section}{Sec.}{Secs.}
\Crefname{section}{Section}{Sections}
\Crefname{table}{Table}{Tables}
\crefname{table}{Tab.}{Tabs.}

\makeatletter
\DeclareRobustCommand\onedot{\futurelet\@let@token\@onedot}
\def\@onedot{\ifx\@let@token.\else.\null\fi\xspace}

\def\etal{\emph{et al}\onedot}
\makeatother

\begin{document}

\ifFGfinal
\thispagestyle{empty}
\pagestyle{empty}
\else
\author{Anonymous FG2024 submission\\ Paper ID \FGPaperID \\}
\pagestyle{plain}
\fi
\maketitle

\begin{abstract}
We address the challenges in estimating 3D human poses from multiple views under occlusion and with limited overlapping views.
We approach multi-view, single-person 3D human pose reconstruction as a regression problem and propose a novel encoder-decoder Transformer architecture to estimate 3D poses from multi-view 2D pose sequences.
The encoder refines 2D skeleton joints detected across different views and times, fusing multi-view and temporal information through global self-attention. 
We enhance the encoder by incorporating a geometry-biased attention mechanism, effectively leveraging geometric relationships between views. 
Additionally, we use detection scores provided by the 2D pose detector to further guide the encoder’s attention based on the reliability of the 2D detections.
The decoder subsequently regresses the 3D pose sequence from these refined tokens, using pre-defined queries for each joint. 
To enhance the generalization of our method to unseen scenes and improve resilience to
missing joints, we implement strategies including scene centering, synthetic views, and
token dropout. 
We conduct extensive experiments on three benchmark public datasets, Human3.6M, CMU Panoptic and Occlusion-Persons. 
Our results demonstrate the efficacy of our approach, particularly in occluded scenes and when few views are available, which are traditionally challenging scenarios for triangulation-based methods.
\end{abstract}


\section{Introduction}
\label{sec:intro}
\input{intro.tex}

\section{Related Work}
\label{sec:related}
\input{related.tex}

\section{Method}
\label{sec:method}
\input{method.tex}

\section{Experiments}
\label{sec:experiments}

\input{experiments.tex}

\section{Conclusion}
\label{sec:conclusion}
\input{conclusion.tex}

{\small
\bibliographystyle{ieee}
\bibliography{egbib}
}

\end{document}

%% file: intro.tex
3D human pose estimation is a fundamental problem in computer vision with applications in many domains, including sports analysis, robotics, virtual reality, and character animation. 
Accurate 3D pose estimation is essential for many of these applications, as it enables understanding human motion, behaviour, and interaction with the environment. 

Although 3D pose estimation methods from monocular images or videos have achieved impressive progress in recent years \cite{martinez_2017_3dbaseline,Moreno_Noguer_2017,Wehrbein2021-vw,Pavllo2019-bz,Rhodin-2018}, estimating human pose in 3D from a single viewpoint is a challenging task, vulnerable to occlusion and inherently ill-posed due to depth ambiguities.

A natural alternative to overcome these limitations is to estimate 3D human poses from multiple camera viewpoints, which enables reconstruction without depth ambiguity and, given calibrated cameras, in absolute world coordinates.
Multi-view 3D pose estimation methods usually operate in two steps: estimating 2D poses in each view, then reconstructing the 3D poses from the 2D poses using for example triangulation \cite{Hartley2003-qk}.
As the reconstruction quality depends greatly on the quality of the 2D poses, many works focus on improving these by fusing 2D features across views \cite{Iskakov2019-jw,Zhang2020-ss,He2020-nn,Remelli2020-kz,Ma2021-gs,Xie2020}.

Most public multi-view human pose datasets are characterized by a central capture area, overlapping views and no occlusions. 
In real applications, however, multi-view camera systems are subject to constraints such as cost, room layout, the presence of furniture or the availability of electrical outlets, resulting in wide-baseline setups with few cameras, strong occlusions, areas where few views overlap and subjects moving in and out of view. 
These are settings in which triangulation-based methods typically struggle. 
Moreover, large rooms or outdoor scenes may require wireless distributed systems in which it is not practical to transmit images or intermediate 2D feature maps for real-time applications.

To address these issues, we approach multi-view, single-person 3D human pose reconstruction as a regression problem and propose a novel encoder-decoder Transformer architecture to estimate 3D poses from multi-view 2D skeleton sequences. 
We treat all 2D joints detected in different views at different times as individual tokens and feed them to the encoder, which fuses multi-view and temporal information using global self-attention.
We propose a geometry-biased attention mechanism to leverage the geometric relationships between different views. 
Additionally, we use the detection scores provided by the 2D pose detector to further guide the encoder’s attention based on the reliability of the 2D detections.
Finally, we regress the 3D poses by querying this set-latent representation of the multi-view pose sequence with a Transformer decoder, using a pre-defined query for each joint.

\begin{figure*}[t!]
\centering
\includegraphics[width=\linewidth]{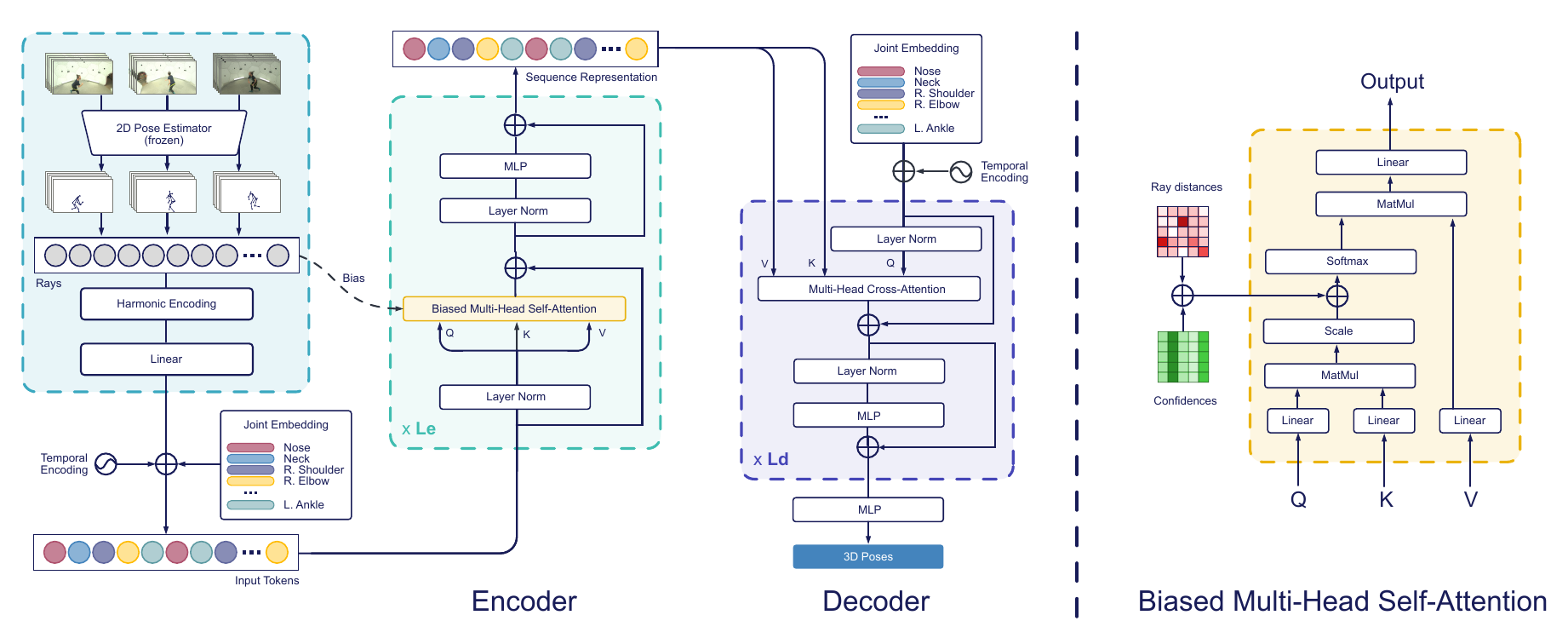}
\caption{\label{fig:arch}\textbf{Overview of our proposed method.}
We obtain 2D pose sequences from multiple views and encode them into a set of joint tokens.
The tokens are processed by a Transformer encoder, yielding a sequence representation via self-attention. The encoder is biased with detection confidence scores and pairwise ray distances (right).
We decode the sequence representation in a Transformer decoder by cross-attention with a set of predefined joint queries.
Finally we regress the 3D coordinates of the joints with an MLP.
 }
\vspace{-5pt}
\end{figure*}

Our main contributions are summarized as follows:
\begin{itemize}
    \item We present a new transformer-based method for multi-view, single-person 3D human pose reconstruction, which effectively fuses multi-view and temporal information.
    \item We introduce confidence and geometry biases to guide the encoder's attention mechanism using the known structure of the problem.
    \item We propose scene centering, synthetic views and token dropout to promote generalization to unseen scenes and make our model resilient to missing joints.
    \item We analyze the performance of our method through ablation studies and show that it outperforms state-of-the-art methods in occluded scenes and when few overlapping views are available.
\end{itemize}

The remainder of this paper is organized as follows: Section \ref{sec:related} discusses previous works related to 3D human pose estimation. 
Section \ref{sec:method} presents the details of our proposed method, including the encoder-decoder Transformer architecture with biased self-attention. 
Section \ref{sec:experiments} reports the experimental results and comparisons with existing methods. 
Finally, Section \ref{sec:conclusion} concludes the paper and discusses possible future directions.

%% file: related.tex
\noindent {\bf 3D Human Pose Estimation.}
Many 3D human pose estimation works focus on monocular images or videos \cite{martinez_2017_3dbaseline,Moreno_Noguer_2017,Wehrbein2021-vw,Pavllo2019-bz,Rhodin-2018}. 
This is however an ill-posed problem due to depth ambiguities and occlusions.
A natural solution to overcome this issue is to estimate 3D poses from multiple views, which is not prone to depth ambiguities.
Moreover, a body part that is occluded in one view may become visible in other views.
Multi-view 3D pose estimation is usually performed in two steps: first extracting 2D keypoints from each view independently with a 2D pose detector, then reconstructing the 3D poses via triangulation \cite{Iskakov2019-jw,Zhang2020-ss,He2020-nn,Remelli2020-kz,Ma2021-gs,Xie2020} or pictorial structure models \cite{Burenius_2013_CVPR,Pavlakos_2017_CVPR,Qiu2019-zi}.
As the accuracy of the reconstruction step greatly depends on the quality of the keypoints obtained at the first step, several recent methods focus on refining the 2D pose detections by fusing information between views.
Qiu \etal \cite{Qiu2019-zi} proposed to learn a fixed attention matrix to retrieve 3D poses from 2D views. However, the fusion network needs to be re-trained if the camera configuration changes.
Iskakov et al. \cite{Iskakov2019-jw} train a 2D detector while solving a multi-view 3D reconstruction problem via differentiable triangulation. They also propose to learn 3D pose directly by fusing features in 3D voxel feature maps. While very accurate, volumetric approaches are computationally expensive.
Several methods focus on refining the detector's intermediate feature maps by leveraging epipolar geometry \cite{Zhang2020-ss,He2020-nn} or a relaxation thereof \cite{Ma2021-gs}.
In particular, AdaFuse \cite{Zhang2020-ss} tackles the occlusion issue by learning an adaptive fusion weight to reduce the impact of low-quality views.
However, it discards information off the epipolar line and treats each joint independently.

\noindent {\bf Transformers for 3D Pose Estimation}
The Transformer \cite{Vaswani2017-si} has proven to be a powerful architecture with
broad applications in various fields, from natural language processing \cite{Vaswani2017-si,edunov-etal-2018-understanding} to computer vision \cite{dosovitskiy2020vit,Carion2020-rf}.
Several researchers have applied Transformers to 2D human pose estimation \cite{mao2022poseur,Xia2022-dl,yang2021transpose}.
Metro \cite{Lin2021-eu} uses a Transformer to directly regress 3D poses and meshes from monocular images.
MixSTE \cite{Zhang_2022_CVPR} performs 3D pose estimation from videos by alternating spatial and temporal attention.

Recently, researchers have used Transformers for multi-view 3D pose estimation.
He \etal proposed the Epipolar Transformer \cite{He2020-nn}, which refines the intermediate features of a 2D detector for a reference view by using features of corresponding points along the epipolar line in neighbouring views.
TransFusion \cite{Ma2021-gs} performs cross-view fusion via global attention. They introduce the epipolar field to encourage high similarity between pixels located close to each other’s epipolar line.
Global attention between 2D feature maps is however computationally expensive.
To alleviate this cost, Ma \etal proposed to prune visual tokens based on the attention scores of keypoint queries \cite{Ma2022-me}.
In this study, we present an encoder-decoder transformer model operating on 2D joint coordinates and associated confidence scores, which is a much sparser input.

%% file: method.tex
\subsection{Overview}
\Cref{fig:arch} provides an overview of our proposed multi-view 3D human pose reconstruction method.
We use an off-the shelf 2D human pose detector \cite{Sun2019-cl} to extract 2D skeletons from a sequence of multi-view images capturing the scene from different viewpoints.
We flatten these poses to a set of joints, and transform them into 3D rays traced through each 2D joint detection and the corresponding camera center.
We project the rays independently to high-dimensional features and refine this set of feature vectors via self-attention in a Transformer encoder, yielding a global set-latent representation for the whole sequence.
We use prior knowledge about geometry and detection scores to guide the attention layers of the encoder.
Finally, we predict the 3D poses with a Transformer decoder by querying the sequence encoding with a set of target feature vectors describing the semantics of the points to decode.

\subsection{Ray Encoding}
Our model takes as input a multi-view sequence of 2D poses with associated confidence scores $\mathbf{S} \in \mathbb{R}^{T_{in} \times C \times J \times 3}$, along with extrinsic camera matrices $\{\mathbf{E}_c=[\mathbf{R}_c|\mathbf{t}_c] \in SE(3)\}_{c=1}^{C}$ and intrinsic  parameters $\{\mathbf{K}_c \in \mathbb{R}^{3 \times 3}\}_{c=1}^{C}$, where $C$ is the number of cameras, $T_{in}$ is the number of time frames and $J$ is the number of keypoints of the detected poses.

We flatten the sequence to a set of individual joints, which we transform using the camera matrices
into 3D rays passing through the estimated 2D joints and the corresponding camera centers.
With perfect camera calibration and perfect detections, these rays would also pass through the corresponding 3D joints.
We use the Plücker representation for the rays \cite{Plucker1865-ab, Jia2020-sg, Sitzmann2021-zf, Venkat2023-zk}, as it decouples the ray coordinates from the positions of the camera centers along the rays, thus promoting generalization to novel camera configurations.
Specifically, for a joint $j$ with pixel coordinates $(u, v)$ in camera $c$, we calculate the Plücker coordinates $\mathbf{r}_{j,c} \in \mathbb{R}^6$ of the ray as
\begin{equation}
	\mathbf{r}_{j,c} = (\mathbf{d}_{j,c}, \mathbf{m}_{j,c}) = (\mathbf{\tilde{d}}_{j,c}, \mathbf{t}_c \times \mathbf{\tilde{d}}_{j,c})/\|\mathbf{\tilde{d}}_{j,c}\|, 
\end{equation}
where $\mathbf{\tilde{d}}_{j,c}$ is the ray direction:
\begin{equation}
    \mathbf{\tilde{d}}_{j,c} = \mathbf{R}^{-1}_c \mathbf{K}_c^{-1}
    \left(
    \begin{smallmatrix}
    u\\
    v\\
    1\\
    \end{smallmatrix}
    \right).
\end{equation}
Finally, we form the ray embeddings as follows:
\begin{equation}
	\mathbf{f}_{j,c} = \mathbf{W}_r(h(\mathbf{r}_{j,c})),
\end{equation}
where $h(\cdot)$ denotes harmonic embedding \cite{Mildenhall2020-wn} and $\mathbf{W}_r$ is a linear projection layer.

\subsection{Biased Transformer Encoder}
We process the ray embeddings using a Transformer encoder \cite{Vaswani2017-si} consisting of $L_e$ layers of global multi-head self-attention.
This allows to model relationships between body joints, across views and along the time dimension simultaneously, providing global context to each token.
As the Transformer is permutation-invariant, we add to the tokens a learned body joint embedding and a harmonic temporal encoding based on the relative timestamps of the observations from the current time step, thus retaining temporal and body structure information.

At the core of the Transformer architecture, the attention mechanism computes a weighted sum of a set of tokens based on their relevance to a particular query.
In particular, scaled dot-product attention calculates the relevance score of each key $\mathbf{K}$ with respect to a query $\mathbf{Q}$:
\begin{equation}
\text{Attention}(\mathbf{Q}, \mathbf{K}, \mathbf{V}) = \operatorname{softmax} \left(\frac{\mathbf{Q}\mathbf{K}^T}{\sqrt{d_k}}\right) \mathbf{V},
\end{equation}
where $\mathbf{Q}$,$\mathbf{K}$, and $\mathbf{V}$ are projections of the input tokens and $d_k$ is the dimension of the feature vectors.

Our problem has known structure, which we can use to guide the transformer during self-attention.
First, the pose detector provides confidence scores indicating the reliability of the predicted coordinates.
Intuitively, when calculating the attention matrix more weight should be given to tokens corresponding to joints with high confidence scores.
To incorporate this knowledge about joint confidence, we bias the attention matrix by adding the confidence matrix $\mathbf{M}_{conf}$, obtained by repeating the row-vector of the keys' detection confidence scores $q$ times row-wise.
For each query, this will increase the attention score of keys corresponding to joints with high confidence scores.

We also posit that a token should attend more to tokens corresponding to the same body joint seen from different cameras and to tokens corresponding to other joints of the body seen from the same viewpoint.
These correspond to rays that are closer in 3D space.
Similarly to \cite{Venkat2023-zk}, we form the matrix $\mathbf{M}_{dist}$ of pairwise distances between the rays, and use it as a geometry bias to the attention map.
Given two rays $\mathbf{r}_{q}=(\mathbf{d}_{q},\mathbf{m}_{q}), \mathbf{r}_{k}=(\mathbf{d}_{k},\mathbf{m}_{k})$ with \textit{unit} directions $\mathbf{d}_{q}$ and  $\mathbf{d}_{k}$, their distance is given by
\begin{equation}
\label{eq:ray_distance}
d (\mathbf{r}_{q}, \mathbf{r}_{k}) = 
\begin{cases}
    \frac{|\mathbf{d}_{q} \cdot \mathbf{m}_{k} + \mathbf{d}_{k} \cdot \mathbf{m}_{q}|}{||\mathbf{d}_{q} \times \mathbf{d}_{k}||},& \mathbf{d}_{q} \times \mathbf{d}_{k} \ne 0 \\[2mm]
    ||\mathbf{d}_{q} \times (\mathbf{m}_{q} - \mathbf{m}_{k})||, & \text{otherwise.}
\end{cases}
\end{equation}
We multiply the distance matrix by a negative factor, learned independently for each encoder layer.
Hence rays with large distance to query rays will be penalized, while intersecting rays, such as rays from the same camera, will have a pairwise distance of 0 and will not be penalized.
Rays corresponding to different views of the same joint at the same time will typically have small pairwise distances, but outliers will be further away from other observations of the same joint and will incur a penalty.
This relates the geometry bias to epipolar geometry, as the attention score will be higher for joint detections along the epipolar line of the query joint.

At each layer $l$ of the encoder, we thus calculate the following biased attention map:
\begin{equation}
\label{eq:attention_weights_conf}
    \mathbf{A}_l = \operatorname{softmax} \Big(\frac{\mathbf{Q}\mathbf{K}^T}{\sqrt{d_k}} + \eta_l^2\mathbf{M}_{conf} - \gamma_l^2\mathbf{M}_{dist}\Big),
\end{equation}
\noindent where $\eta_l$ and $\gamma_l$ are learnable parameters for each layer.

\subsection{3D pose sequence decoding}
The output of the encoder is a set of feature vectors representing the whole pose sequence.
To extract information about specific joints at given times, we construct a predefined query providing the semantics of the points to decode.
It consists in $J$ learned embedding vectors corresponding to each human keypoint, repeated $T_{out}$ times, where $T_{out}$ is the desired time range.
We add harmonic temporal encoding to specify the timing of each point.
We decode the 3D pose sequence using a Transformer decoder, 
which attends to the set of feature vectors with several layers of multi-head cross-attention with the predefined query.
Finally, we regress the 3D world coordinates of the joints from the decoded tokens using a Multi-Layer Perceptron (MLP).

The query does not need to comprise the same number of frames as the input\footnote{In fact, the query does not even need to contain the same number of joints as the input, although this is not something we explore here.}.
Moreover, the number of input and output frames can be chosen to be different during training and inference.

\subsection{Training and Inference}

The model is trained by minimizing the Mean Squared Error (MSE)
\begin{equation}
	\mathcal{L} = \frac{1}{N T_{out} J} \sum_{i=1}^{N} \sum_{t=1}^{T_{out}} \sum_{j=1}^{J} \|P_{itj} - \hat{P}_{itj}\|^2,
\end{equation}
where N is the number of poses in the dataset, $P_{itj}$ is a ground-truth 3D joint and $\hat{P}_{itj}$ is the corresponding prediction.

We train our model to reconstruct the whole sequence of length $T_{out} = T_{in}$. 
During inference, we input $T_{in}$ frames of observations, i.e. the latest observation and $T_{in} - 1$ past observations, but we are only interested in reconstructing the latest frame, i.e. $T_{out}=1$, regardless of $T_{in}$.
Hence our model can make fully causal predictions and is well-suited for real-time inference.

\subsection{Promoting Generalization}
Multi-view pose datasets with 3D annotations are relatively small, and are usually collected in controlled environments with fixed capture area, few camera poses and no occlusions.
To enable our model to handle unseen scenes and camera configurations, we introduce the following methods during training and inference.

\noindent {\bf Centering.}
To enable the model to generalize to scenes of arbitrary dimensions, we \textit{roughly} center the scene around the subject during training and inference.
During training, given an input sequence of length $T_{in}$, we select a random ground-truth pose at time $t$ and project its neck joint on the floor. 
We add random noise on the plane to this projection, and transform all observation rays and ground-truth poses of the sequence to a coordinate system centered on the resulting point.
Finally, we rotate the scene around the vertical axis with a random angle.
During inference, at the first time step the 3D rays can be centered around a point obtained by triangulation of the observations, without rotating the scene.
At subsequent time steps, the coordinate system is centered around the predicted 3D pose obtained at the previous time step.

As shown in \cref{sec:ablations}, transforming the scene to this body-centric coordinate system allows our model to better generalize to scenes with unseen dimensions.

\begin{table*}[hbt]
\begin{minipage}{\columnwidth}
\caption{\label{table:fewer-cams}Reconstruction error on H36M dataset with fewer cameras. Absolute MPJPE (mm). "$\dagger$": Pose detector fine-tuned on H36M.} 
\small
\centering
\setlength{\tabcolsep}{4pt} 
\begin{tabular*}{\columnwidth}{@{\extracolsep{\fill}}ll ccc@{}} 
\toprule[1pt]
 & & \multicolumn{3}{c}{Number of Cameras}\B\\
Method & & 2 & 3  & 4 \\
\midrule[.6pt]
Alg. Tri. \cite{Iskakov2019-jw} & (ResNet-152)$\dagger$ &  \hphantom{1}51.1 &  \textbf{23.4} & \textbf{19.1} \\
Ours & (ResNet-152)$\dagger$ &  \hphantom{1}\textbf{29.9} &  24.4 & 22.7 \\
\midrule[.5pt]
Alg. Tri. &(HRNet) &  120.7 &  50.9 & 44.2 \\
Ours & (HRNet) &  \hphantom{1}\textbf{36.8} &  \textbf{30.4} & \textbf{26.0} \\
\bottomrule[1pt]
\end{tabular*}
\end{minipage}\hfill 
\begin{minipage}{\columnwidth}
\caption{\label{table:h36m-occl}Reconstruction error on the H36M-Occl test dataset. Absolute MPJPE (mm). "$\dagger$": Pose detector fine-tuned on H36M.}
\small
\centering
\setlength{\tabcolsep}{4pt} 
\begin{tabular*}{\columnwidth}{@{\extracolsep{\fill}}ll ccc@{}} 
\toprule[1pt]
 & & \multicolumn{3}{c}{Number of Cameras}\B\\
Method & & 2 & 3  & 4 \\
\midrule[.6pt]
Alg. Tri. \cite{Iskakov2019-jw} & (ResNet-152)$\dagger$ &  163.3 &  39.5 & \textbf{27.9} \\
Ours & (ResNet-152)$\dagger$ &  \hphantom{1}\textbf{39.1} &  \textbf{33.4} & 31.3 \\
\midrule[.5pt]
Alg. Tri. & (HRNet) &  217.3 &  72.4 & 54.1 \\
Ours & (HRNet) &  \hphantom{1}\textbf{42.3} & \textbf{34.5} & \textbf{31.6} \\
\bottomrule[1pt]
\end{tabular*}
  \end{minipage}
\end{table*}

\begin{table}
    \small
    \centering
    \caption{\label{table:occl-person}Reconstruction error on Occlusion-Person. Absolute MPJPE (mm).  "$\dagger$": Pose detector fine-tuned on the dataset.}
    \setlength{\tabcolsep}{4pt} 
    \begin{tabular*}{\columnwidth}{@{\extracolsep{\fill}}l cccc@{\quad}} 
    \toprule[1pt]
     & \multicolumn{4}{c}{Number of Cameras}\B\\
    Method & 2 & 3  & 4 & 8\B\\
    \midrule[.6pt]
    RANSAC$\dagger$ \cite{Zhang2020-ss} & 33.7 & 87.5 & 35.0 & 15.5\\
    ScoreFuse$\dagger$ \cite{Zhang2020-ss} & 32.7 & 25.7 & 21.4 & 15.0\\
    AdaFuse$\dagger$ \cite{Zhang2020-ss} & - & 26.2 & 19.7 & \textbf{12.6}\B\\
    \midrule[.5pt]
    Ours (HRNet) & \textbf{30.8} & \textbf{22.9} & \textbf{19.6} & 14.2\T\B\\ 
    \bottomrule[1pt]
    \end{tabular*}
\end{table}    

\noindent {\bf Synthetic views.}
Existing multi-view datasets with ground-truth 3D pose annotations have fixed, limited camera pose configurations.
This can limit the generalization capabilities of learning-based pose reconstruction methods.
One way to mitigate this issue is through the use of synthetic data.
During training, we generate synthetic 2D poses in addition to the observations provided by the datasets.
Specifically, we sample random camera positions and calculate the rays passing through these and each joint of the ground-truth 3D poses.

\noindent {\bf Token dropout.}
To make our model resilient to missing data, we randomly remove 20\% of the input tokens during training, employing a similar technique as \cite{Liu2022PatchDropoutEV}.
As a consequence, a given joint may be invisible in all views during the whole sequence, or the whole body may be missing at given times.
This further encourages the model to aggregate information from the whole body and across time frames.
Token dropout has the convenient side-effect of reducing computing and memory requirements during training.

%% file: experiments.tex
We evaluate our method on three large-scale multi-view datasets for multi-view 3D human pose estimation: Human3.6M \cite{Ionescu2014-ps},  CMU Panoptic \cite{Joo2016-va} and Occlusion-Person \cite{Zhang2020-ss}.
We inspect our method's ability to handle occlusions and scenes with few views, and test its ability to generalize to new datasets.
We also compare our results to the state of the art and conduct ablation studies to validate the effectiveness of the components of our method.

\subsection{Datasets and Evaluation Metrics}

\noindent {\bf Human 3.6M}
\cite{Ionescu2014-ps} consists of 3.6 million images of 11 subjects performing various activities in an indoor environment, with corresponding 3D poses. 
The images are captured at 50 Hz by 4 synchronized cameras placed approximately at the corners of a 4m $\times$ 10m rectangle.
The ground truth 3D pose annotations are obtained by a marker-based motion capture system.
We follow standard practice and use 5 subjects for training (S1, S5, S6, S7, S8) and 2 subjects for testing (S9, S11).
Some sequences with the 'S9' validation subject have erroneous 3D annotations and were ignored during evaluation as in \cite{Iskakov2019-jw}: parts of ’Greeting’, ’SittingDown’ and ’Waiting’.
To test the performance of our method under occlusion, we adopt the method of \cite{Sarandi2018-ft,Zhang2020-si} to create a test dataset derived from H36M, \textit{H36M-Occl}, in which we simulate occlusion scenarios by randomly placing white square masks over the 2D joint, with a probability of $0.1$. 

\noindent {\bf CMU Panoptic}
\cite{Joo2016-va} contains multiple scenes in which people perform different activities in an indoor environment. 
Images are captured by hundreds of VGA cameras and 31 HD cameras installed on a hemispherical structure of diameter 5.5m.
The 3D pose annotations are obtained by triangulation using all camera views.
We use the same train/test sequences as Iskakov \etal \cite{Iskakov2019-jw}, consisting of scenes featuring a single person.
Unless otherwise stated, we use 4 HD cameras (2,13,10,19) for testing and the remaining 27 HD cameras for training.

\noindent {\bf Occlusion-Person}
\cite{Zhang2020-ss} is a synthetic dataset rendered with UnrealCV \cite{Qiu2017-il} in which thirteen human models driven by motion sequences from the CMU Motion Capture database \cite{CMU_2003-gw} are put into nine scenes featuring occluding objects and captured by 8 cameras.
We use the same train/test sequences as Zhang \etal \cite{Zhang2020-ss}.

\noindent {\bf Evaluation metric.}
We evaluate the 3D pose reconstruction accuracy using the absolute Mean Per Joint Position Error (MPJPE), which is the average of the L2 distance between the prediction and the ground truth for each joint, measured in millimeters.
We do not align the estimated 3D poses and ground truth in any way.

\subsection{Implementation Details}
We undistort the images using the camera parameters provided with the datasets and extract the 2D poses with a pre-trained pose detector which remains frozen when training our network.
Unless otherwise indicated, we use HRNet-W32 \cite{Sun2019-cl}, trained on COCO, with YoloX \cite{Ge2021-cq} box proposals.
For the harmonic embedding $h$,  we  use 15 frequencies.
We use 3 layers for the transformer encoder and 2 layers for the decoder. Both encoder and decoder have 6 heads.
During training, we input 9 time frames and 2 randomly sampled views, and also output 9 frames.
Unless otherwise stated, during evaluation we input 9 time frames and only output 1 frame, corresponding to the latest observation.

We train our model with the Adam optimizer \cite{Kingma2014-uu} for 300000 iterations with a mini-batch size of 256.
The initial learning rate is $10^{-4}$. It is smoothly decayed after a warmup phase of 10 k steps.
Training takes about 5 hours on a single Nvidia RTX 4090.

\begin{table*}[t]
\centering
\caption{\label{table:pano-h36m} Generalization between datasets. We train our model on the CMU Panoptic dataset and test on H36M. Absolute MPJPE (mm).}
\resizebox{1.0\textwidth}{!}{
\small
\begin{tabular}{ll ccccccccccccccc|c}
\toprule
\multicolumn{2}{l}{Method} & Dir & Disc & Eat & Greet & Phone & Pose & Purch & Sit & SitD & Smoke & Photo & Wait & Walk & WalkD & WalkT & Avg \\
\midrule
Alg. Tri. &(HRNet) & 44.6 & 43.5 & 42.2 & 42.5 & 43.7 & 41.6 & 48.5 & \textbf{46.8} & \textbf{51.7} & 45.4 & 45.7 & 42.3 & 40.3 & 45.6 & 39.8 & 44.3\\
Ours &(HRNet) & \textbf{33.9} & \textbf{32.5} & \textbf{32.6} & \textbf{33.7} & \textbf{38.3} & \textbf{30.8} & \textbf{35.1} & 56.9 & 85.9 & \textbf{36.0} & \textbf{37.9} & \textbf{33.3} & \textbf{29.1} & \textbf{38.7} & \textbf{28.7} & \textbf{38.9} \\
\bottomrule
\end{tabular}
}
\end{table*}

\subsection{Impact of Varying Number of Views}
We measure the performance of our method on the H36M dataset when fewer camera views are available.
To compare with the Algebraic Triangulation method in \cite{Iskakov2019-jw}, we train and test our network on 2D poses provided by the ResNet-152 detector from Iskakov \etal.
We also compare our method to the Algebraic Triangulation method applied to HRNet poses.
We test all methods with different numbers of input views, and present the average error obtained with all combinations of views. 
The results are shown in \cref{table:fewer-cams}.
With the high-quality ResNet-152 poses, Algebraic Triangulation yields superior results with 3 and 4 views, but the reconstruction accuracy drops considerably when only 2 views are available, possibly due to self-occlusions.
On the other hand, our method's accuracy does not drop as much when using only 2 views.
When using poses extracted with the off-the shelf detector, our method consistently achieves lower reconstruction error regardless of the number of views.

\subsection{Occlusion handling} 
We evaluate our method’s ability to handle occlusions on the H36M-Occl test dataset, using the model trained on the original H36M training set without occlusions.
As shown in \cref{table:h36m-occl}, when using the ResNet-152 detector, fine-tuned on H36M, triangulation yields the best results when enough cameras are available.
However, with fewer than four cameras our method provides significantly better results.
Also, when using the off-the-shelf HRNet pose detector, our method yields superior results compared to the baseline with any number of cameras.

We also compare our method to Adafuse \cite{Zhang2020-ss} and its baselines on the Occlusion-Person dataset.
For all methods in \cite{Zhang2020-ss}, 3D joints are reconstructed independently, using triangulation, hence we only calculate the error for joints visible in at least 2 views.
For our method, on the other hand, we present the error for all joints, including joints that are not visible in any view.
As shown in \cref{table:occl-person}, our method achieves competitive results when all camera views are used.
In this setting, however, most joints are visible in at least 4 views.
Therefore, we also compare the methods with different numbers $N_c$ of input views, and present the average error obtained with all combinations of $N_c$ cameras.
When the number of available views decreases, our method yields the smallest reconstruction errors, further highlighting its ability to handle occluded scenes. 

\begin{table}
\caption{\label{table:h36m} Comparison with state-of-the-art methods. Absolute MPJPE (mm) on the Human3.6M dataset. "$\dagger$": 2D pose detector fine-tuned on the target dataset.}
\small
\centering
\setlength{\tabcolsep}{4pt} 
\begin{tabular*}{\columnwidth}{@{\extracolsep{\fill}}l cc@{\qquad}} 
\toprule[1pt]
\multicolumn{1}{l}{Method} & H36M & CMU \T\B\\
\midrule[.6pt]
 Cross View Fusion \cite{Qiu2019-zi}$\dagger$ &  26.2 & - \T\\
 RANSAC Baseline \cite{Iskakov2019-jw}   & - & 33.4\\ 
 Algebraic Triangulation \cite{Iskakov2019-jw}$\dagger$   & 19.2 & 21.3 \\
 Volumetric \cite{Iskakov2019-jw}$\dagger$   & {\bf 17.7} & {\bf 13.7} \\
 Remelli et al. \cite{Remelli2020-kz}$\dagger$  & 30.2 & - \\
 Epipolar Transformers \cite{He2020-nn}$\dagger$ & 27.1 & - \\
 TransFusion \cite{Ma2021-gs}$\dagger$ & 25.8 & - \\
 Ours (ResNet-152)$\dagger$ & 22.7 & -\T\B\\
\midrule[.4pt]
 Algebraic Triangulation (HRNet) & 44.2& 26.2\T\\ 
 Ours (HRNet)   & \textbf{26.0} & \textbf{17.2}\B\\
\bottomrule[1pt]
\end{tabular*}
\end{table}

\subsection{Generalization Between Datasets} 

To examine if our proposed approach generalizes between datasets, we train our model on the CMU Panoptic dataset and test it on H36M.
Following \cite{Iskakov2019-jw}, we calculate the error on joints that have similar 3D annotations in both datasets (shoulder, elbow, wrist, knee, ankle).
As can be seen in \cref{table:pano-h36m}, our method yields a lower error than the baseline, demonstrating the ability of our model to generalize to unseen camera settings and positions.
However, the results also show that our model struggles to reconstruct correct poses for the sitting sequences. 
These sequences contain body poses that are not present in the CMU Panoptic dataset, which highlights a limitation of our method.

\subsection{Comparison with other methods}
In \cref{table:h36m}, we compare our method with previous work on the H36M and CMU Panoptic datasets.
Other methods use 2D pose detectors fine-tuned on the target dataset. 
To provide a fair comparison, we train our model on 2D poses obtained with the detector from \cite{Iskakov2019-jw}.
In this setting, it achieves competitive results.
To consider a more realistic scenario, we use off-the-shelf algorithms to detect human bounding boxes and extract 2D poses.
As we are not aware of other methods providing absolute pose reconstruction results for off-the-shelf detectors, we apply the Algebraic Triangulation method from \cite{Iskakov2019-jw} to the HRNet detections and compare to our method.
Here, our approach achieves superior results compared to the baseline.

\subsection{Ablation Studies}
\label{sec:ablations}
We evaluate the usefulness of the components of our method by removing them in turn.
For each ablation, we present results with various combinations of training and test datasets.
As seen in \cref{table:ablation}, the reconstruction accuracy degrades when the geometry and confidence biases are removed.
When conducting intra-dataset evaluations, the advantages of synthetic views and centering are unclear; however, they significantly enhance accuracy when generalizing between datasets.
This is likely due to the significant difference in size between the capture areas of the H36M and CMU datasets.

We also study the effect of the number of input time frames.
We report the results of our method on H36M and H36M-Occl, using 4 views and HRNet 2D detections as input.
We use the same model trained on H36M.
As shown in \cref{table:abl-timesteps}, increasing the input sequence length improves reconstruction accuracy.
Temporal information is particularly significant under occlusion.
Even providing only one previous time frame of observations has a strong effect.

\begin{table}
\caption{\label{table:ablation}\textbf{Ablation study} on the components of our framework. We remove each feature in turn and train on the CMU Panoptic and H36M datasets. We report the 3D pose estimation error MPJPE (mm).}
\centering
\resizebox{1.0\columnwidth}{!}{
\small
\begin{tabular}{ll cccccc}
\toprule[1pt]
\multicolumn{2}{l}{Centering}    &      -       & $\checkmark$ & $\checkmark$ & $\checkmark$ & $\checkmark$ & $\checkmark$ \\
\multicolumn{2}{l}{Synthetic} &      -       &      -       & $\checkmark$ & $\checkmark$ & $\checkmark$ & $\checkmark$ \\
\multicolumn{2}{l}{Confidence bias}      &      -       &      -       &      -       & $\checkmark$ &      -       & $\checkmark$ \\
\multicolumn{2}{l}{Geometry bias}  &      -       &      -       &      -       &       -      & $\checkmark$ & $\checkmark$ \\
\midrule
Train & Test\T\\
\cmidrule{1-2}
CMU & CMU & \hphantom{1}22.4 & 18.9 & 19.3 & 17.7 & \textbf{17.1} & 17.2\\ 
H36M & H36M   & \hphantom{1}39.0 & 49.2 & 40.6 & 33.2 & 33.1 & \textbf{26.0}\\
H36M & H36M-Occl & \hphantom{1}43.1 & 52.3 & 44.0 & 35.5 & 37.2 & \textbf{31.6}\\
CMU & H36M & 101.2 & 57.2 & 55.7 & 43.0 & 43.1 & \textbf{38.9}\B\\ 
\bottomrule[1pt]
\end{tabular}
}
\end{table}

\begin{table}
    \small
    \centering
    \caption{\label{table:abl-timesteps}Effect of the number of time frames. Absolute MPJPE (mm). We evaluate the same model trained on H36M with HRNet detections, using 4 camera views.}
    \setlength{\tabcolsep}{4pt} 
    \begin{tabular*}{\columnwidth}{@{\extracolsep{\fill}}l ccccc@{\quad}} 
    \toprule[1pt]
     & \multicolumn{4}{c}{Number of Time Frames}\B\\
    Dataset & 1 & 2 & 3  & 6 & 9\B\\
    \midrule[.6pt]
    H36M & 29.4 & 27.9 & 27.7 & 27.3 & \textbf{26.0}\T\\ 
    H36M-Occl & 41.5 & 34.9 & 34.0 & 32.5 & \textbf{31.6}\B\\ 
    \bottomrule[1pt]
    \end{tabular*}
\end{table}

%% file: conclusion.tex
We presented a novel approach for 3D human pose reconstruction using 2D pose data from multiple viewpoints, leveraging a Transformer-based encoder-decoder architecture with biased self-attention. 
It can straightforwardly accommodate any 2D pose detection network and is suitable for wide-baseline, real-time systems.
Experiments on three large-scale datasets show that our method consistently outperforms triangulation-based methods under heavy occlusion and when fewer input views are available.
One of the major limitations of our approach is its difficulty in reconstructing unseen human poses. 
One potential strategy to mitigate this challenge it to pre-train the network on large-scale 3D motion capture datasets such as \cite{AMASS:ICCV:2019}.
There may also be a benefit in exploring self-supervised training, which would enable our network to be trained on substantially larger datasets without 3D annotations.
Further interesting avenues for future work include improving the model's performance by decoupling the size of the latent set from the number of the inputs in a way similar to \cite{jaegle2022perceiver}, and extending our approach to accommodate multi-person settings.